\documentclass[letterpaper, 10 pt, conference]{ieeeconf}  

\IEEEoverridecommandlockouts                              

\makeatletter
\let\NAT@parse\undefined
\makeatother
\usepackage[numbers]{natbib}

\usepackage[pdftex]{graphicx}
\usepackage{amsmath}
\usepackage{amssymb}  
\usepackage{subfig}
\usepackage{multirow}
\usepackage{array,booktabs}
\usepackage{diagbox}
\usepackage{balance}
\usepackage[ruled]{algorithm}
\usepackage{algpseudocode}
 \usepackage[final]{hyperref}
 \hypersetup{
     colorlinks=true,
     linkcolor=blue,
     filecolor=magenta,
     urlcolor=blue,
     citecolor=black
 }
\usepackage[font=small]{caption}
\usepackage{./rpm_packages/rpm_SIunits}
\usepackage{./rpm_packages/rpm_acronyms}
\usepackage{./rpm_packages/rpm_math}
\usepackage{./rpm_packages/rpm_misc}

\usepackage{soul,color}
\usepackage{lipsum}

\usepackage{xcolor}

\usepackage{tikz} 

\title{\LARGE \bf
ConPR: Ongoing Construction Site Dataset for Place Recognition
}     

\author{Dongjae Lee${}^{1}$, Minwoo Jung${}^{1}$, and Ayoung Kim${}^{1*}$
\thanks{$^\dagger$This work is supported by the Korea Agency for Infrastructure Technology Advancement (KAIA) grant funded by the Ministry of Land, Infrastructure and Transport (Grant RS-2023-00250727) through the Korea Floating Infrastructure Research Center at Seoul National University.}%
\thanks{$^{1}$D. Lee, M. Jung and A. Kim are with the Department of Mechanical Engineering, SNU, Seoul, S. Korea {\tt\small [pur22, moonshot ayoungk]@snu.ac.kr}}%
}

\begin{document}

\makeatletter
  \let\@oldmaketitle\@maketitle
  \renewcommand{\@maketitle}{\@oldmaketitle
  \bigskip
  \centering
    \includegraphics[width=0.95\textwidth]{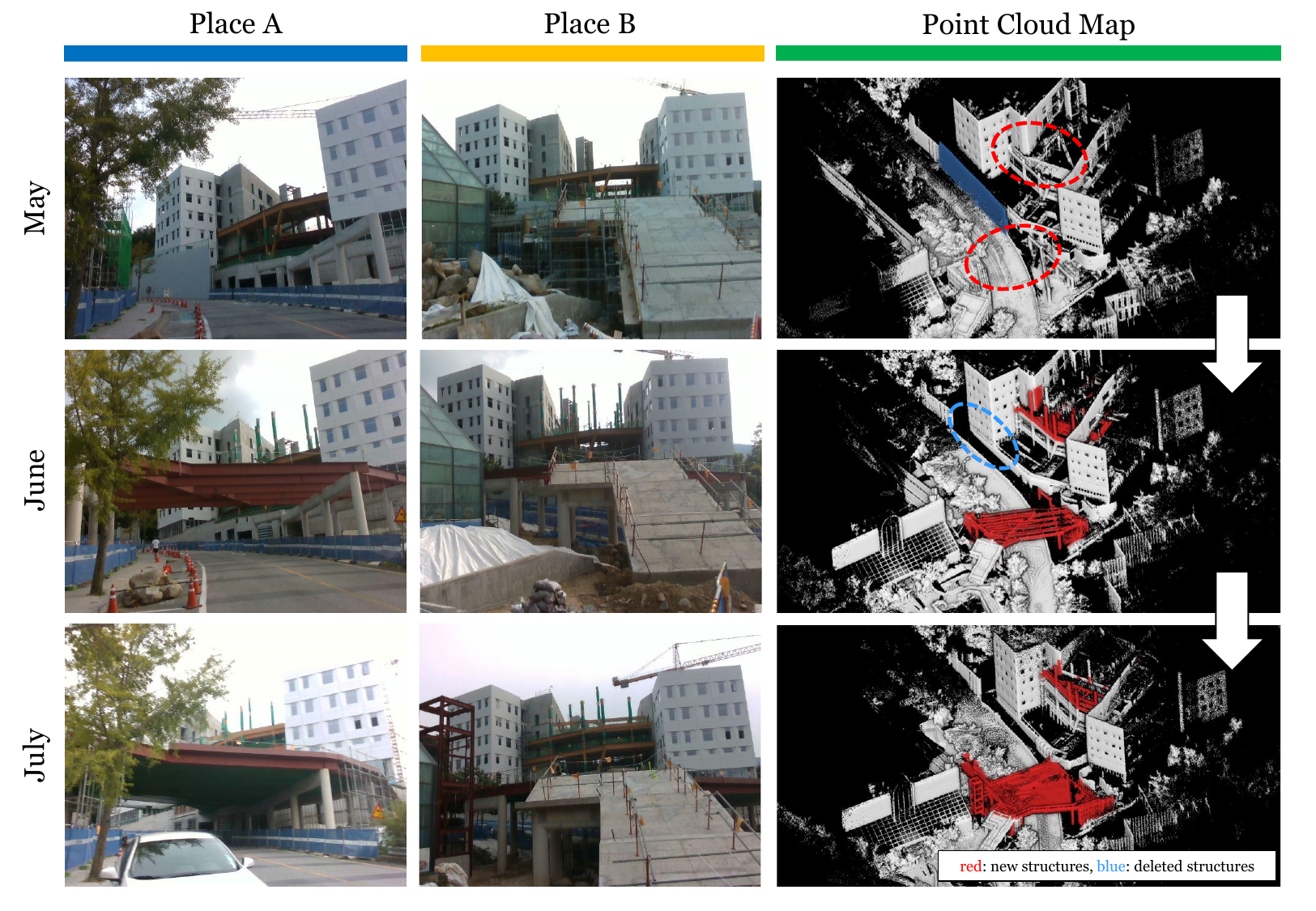}
    \captionof{figure}{
      These sample RGB images and 3D point cloud maps are extracted from our ongoing construction site dataset. The dataset captures the development of buildings and bridges in the active construction site. In addition to showcasing the construction progress, this dataset presents various challenges, including the presence of dynamic objects like vehicles, pedestrians, and construction equipment, along with variations in lighting conditions and viewpoints.
    }
    \vspace{-3mm}
    \label{fig:main}
  }

\makeatother

\maketitle

\renewcommand{\thefigure}{\arabic{figure}}
\setcounter{figure}{1}

\thispagestyle{empty}
\pagestyle{empty}

\begin{abstract}

Place recognition, an essential challenge in computer vision and robotics, involves identifying previously visited locations. Despite algorithmic progress, challenges related to appearance change persist, with existing datasets often focusing on seasonal and weather variations but overlooking terrain changes. Understanding terrain alterations becomes critical for effective place recognition, given the aging infrastructure and ongoing city repairs. For real-world applicability, the comprehensive evaluation of algorithms must consider spatial dynamics. To address existing limitations, we present a novel multi-session place recognition dataset acquired from an active construction site. Our dataset captures ongoing construction progress through multiple data collections, facilitating evaluation in dynamic environments. It includes camera images, LiDAR point cloud data, and IMU data, enabling visual and LiDAR-based place recognition techniques, and supporting sensor fusion. Additionally, we provide ground truth information for range-based place recognition evaluation. Our dataset aims to advance place recognition algorithms in challenging and dynamic settings. Our dataset is available at \href{https://github.com/dongjae0107/ConPR}{https://github.com/dongjae0107/ConPR}.

\end{abstract}
\section{introduction \& related works}










 



Place recognition stands as a foundational task in the fields of computer vision and robotics, with the objective of discerning and establishing correspondences between previously encountered sites. It plays a crucial role in various applications, such as robot localization and navigation. Initially, place recognition algorithms were predominantly developed for visual sensors \cite{galvez2012bags, arandjelovic2016netvlad}, but there has been a growing interest in and active development of algorithms utilizing LiDAR sensors as well \cite{uy2018pointnetvlad, dube2020segmap, kim2021scan}.

Despite notable advancements in place recognition algorithms, several challenges remain, particularly in handling appearance change. Appearance change results from dynamic objects, lighting variations, weather fluctuations, and landscape alterations. While numerous datasets address appearance change \cite{maddern20171, chen2017only, olid2018single, warburg2020mapillary}, most of them mainly focus on seasonal, weather, and time variations, with limited emphasis on changes in terrain. However, many locations undergo frequent repair and construction activities, resulting in substantial alterations to the surrounding terrain and appearance. Therefore, to ensure the applicability of place recognition algorithms in real-world scenarios where construction and urban development are ongoing processes, a comprehensive evaluation must encompass terrain changes. 

AmsterTime \cite{yildiz2022amstertime} provides historical archival images and current images for place recognition, capturing spatial changes over time. However, it is not ideal for places experiencing frequent changes, such as construction sites, due to the significant time gaps between image sets.

ConSLAM \cite{trzeciak2023conslamExtension}, offers real-world construction data collected over 4 months on an active site, making it suitable for the challenge we highlighted. However, ConSLAM primarily focuses on testing odometry and SLAM algorithms in indoor settings, which poses a significant difference from our dataset, which is specifically designed for place recognition in outdoor environments.

In this paper, we introduce a new dataset acquired during the construction period from the vicinity of an active construction site, specifically designed for multi-session place recognition. Our dataset offers several key contributions:
\begin{itemize}
    \item Spatial Dynamics: Our dataset comprises multiple data collections from the surrounding area of the construction site, capturing the ongoing progress of the construction process. This enables the evaluation of place recognition algorithms in dynamic environments where terrain changes occur over time.
    \item Sensor Modalities: In addition to camera images, we provide LiDAR point cloud data. This dual-sensor dataset enables the utilization of both visual and LiDAR-based place recognition techniques. Moreover, it facilitates sensor fusion to enhance place recognition performance and enables cross-modal place recognition for diverse applications.
    \item Ground Truth for Evaluation: We provide ground truth information for range-based place recognition evaluation. This allows researchers to quantitatively analyze the accuracy and robustness of algorithms relying on range measurements.
\end{itemize}

\section{System Overview}
\subsection{Sensor Configuration}
\begin{figure}[t]
    \vspace{-1mm}
	\centering
		\includegraphics[width=0.85\columnwidth]{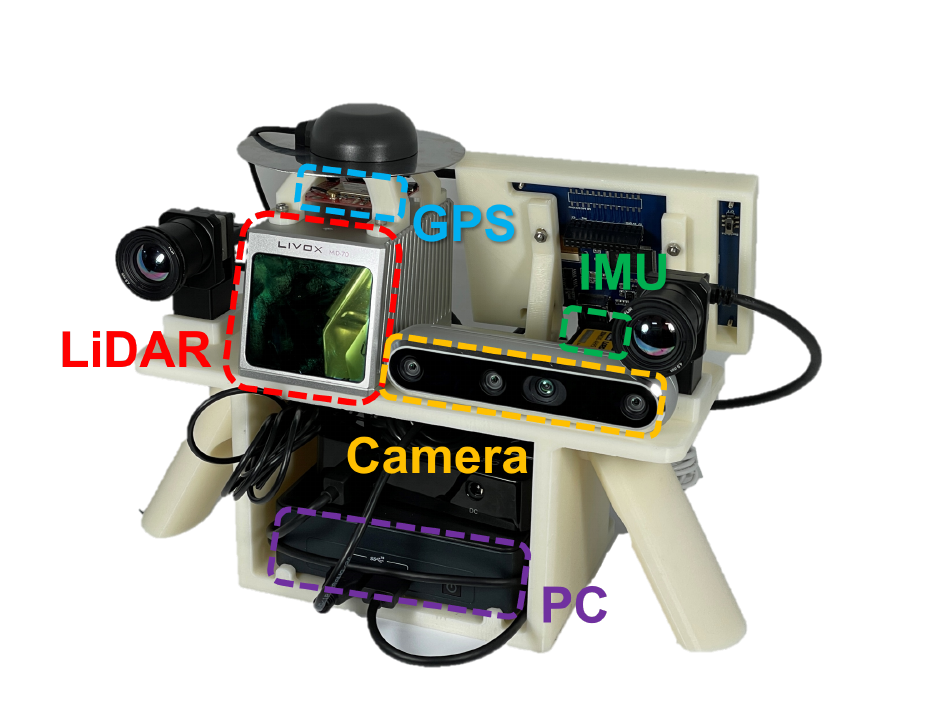}
  	\caption{Handheld system incorporating a monocular camera and solid-state LiDAR.}
	\label{fig:sensor configuration}
	\vspace{-6mm}
\end{figure}

To facilitate unrestricted data acquisition near the construction site, we equipped a handheld system with a monocular camera and a 3D solid-state LiDAR, as in \figref{fig:sensor configuration}. In addition, we have also attached an IMU and GPS to the handheld system to enable its utilization for other purposes (e.g., visual-inertial odometry, LiDAR-inertial odometry).

\subsection{Sensor Calibration}
We performed intrinsic calibration for the RGB camera using a checkerboard pattern, and the overall mean error of the reprojection error is 0.22 pixels. For extrinsic calibration between the LiDAR and IMU, we utilized the method proposed in \cite{zhu2022robust}. Extrinsic calibration between the camera and IMU was achieved using the method proposed in \cite{6696514}.
\section{Ongoing Construction Site Dataset}

Our dataset revolves around a campus environment, focusing on an active construction site responsible for building a new structure and a bridge connecting to an existing building. It consists of 12 sequences, each capturing the construction progress on different dates. While there are no revisits or reverse revisits in any single sequence, the movement paths remain nearly identical across all sequences. This design ensures the dataset's suitability for handling the challenges of recognizing places across multiple sessions. The movement paths traverse diverse environments, including driveways, stairs, and short narrow forest roads, providing varied scenarios for evaluating place recognition performance.

The dataset offers valuable insights into changes in the surrounding landscape and the presence of dynamic elements such as cranes, workers, and pedestrians. Data collection was conducted using a handheld system, which introduced challenging factors like motion blur and varying viewpoints. Notably, the dataset includes sequences captured at night, facilitating research on illumination change. To enable range-based place recognition evaluation, the ground truth positions for all sequences were acquired via FAST-LIO2 \cite{9697912} with GPS.

\section{conclusion}

In conclusion, our paper introduces a novel dataset designed for multi-session place recognition in dynamic construction environments. With multiple data collections capturing ongoing changes in an active construction site, our dataset allows for the evaluation of place recognition algorithms in dynamic environments. We hope that our dataset will serve as a valuable benchmark for advancing place recognition research in diverse and evolving environments.


\balance
\small
\bibliographystyle{IEEEtranN} 
\bibliography{string-short,references}

\end{document}